%% file: final_iclr2026_snippet_gait.tex
\documentclass{article} 
\usepackage{iclr2026_conference,times}

\input{math_commands.tex}

\usepackage{hyperref}
\usepackage{url}
\usepackage{enumerate}
\usepackage{subfigure}
\usepackage{multirow}
\usepackage{amsthm,amsmath,amssymb}
\usepackage{mathrsfs}
\usepackage{rotating}
\usepackage{xcolor}
\usepackage{bm}
\usepackage{wrapfig}
\usepackage[dvipsnames,svgnames,x11names]{xcolor}

\newcommand{\bftab}[1]{{\fontseries{b}\selectfont#1}}
\newcommand{\tabincell}[2]{\begin{tabular}{@{}#1@{}}#2\end{tabular}}
\newcommand{\etal}{\textit{et al}}
\newcommand{\etc}{\textit{etc}}
\newcommand{\ie}{\textit{i.e.}}
\newcommand{\eg}{\textit{e.g.}}

\newcommand{\RED}[1]{\textcolor{BrickRed}{\bftab{#1}}}
\newcommand{\BLUE}[1]{\textcolor{NavyBlue}{\bftab{#1}}}

\title{GaitSnippet: Gait Recognition Beyond Unordered Sets and Ordered Sequences}

\author{Saihui Hou$^{1}$, Chenye Wang$^{1}$, Wenpeng Lang$^{1}$, Zhengxiang Lan$^{1}$, \& Yongzhen Huang$^{1,2}$\thanks{Corresponding author.} \\
$^1$School of Artificial Intelligence, Beijing Normal University $^2$WATRIX.AI \\
\texttt{\{housaihui, huangyongzhen\}@bnu.edu.cn} \\
\texttt{\{chenye.wang, wenpenglang, zhengxianglan\}@mail.bnu.edu.cn} \\
}

\iclrfinalcopy 
\begin{document}

\maketitle

\input{0_abstract}
\input{1_introduction}
\input{2_related_work}
\input{3_approach}

\input{4_experiments}
\input{5_conclusion}

\bibliography{6_reference}
\bibliographystyle{iclr2026_conference}

\appendix
\input{X_supp}

\end{document}

%% file: math_commands.tex
\usepackage{amsmath,amsfonts,bm}

\def\eqref#1{equation~\ref{#1}}

\def\1{\bm{1}}

\DeclareMathAlphabet{\mathsfit}{\encodingdefault}{\sfdefault}{m}{sl}
\SetMathAlphabet{\mathsfit}{bold}{\encodingdefault}{\sfdefault}{bx}{n}



%% file: 0_abstract.tex

\begin{abstract}
Recent advancements in gait recognition have significantly enhanced performance by treating silhouettes as either an unordered set or an ordered sequence.
However, both set-based and sequence-based approaches exhibit notable limitations.
Specifically, set-based methods tend to overlook short-range temporal context for individual frames, while sequence-based methods struggle to capture long-range temporal dependencies effectively.
To address these challenges, we draw inspiration from human identification and propose a new perspective that conceptualizes human gait as a composition of individualized actions.
Each action is represented by a series of frames, randomly selected from a continuous segment of the sequence, which we term a \textbf{snippet}.
Fundamentally, the collection of snippets for a given sequence enables the incorporation of multi-scale temporal context, facilitating more comprehensive gait feature learning.
Moreover, we introduce a non-trivial solution for snippet-based gait recognition, focusing on Snippet Sampling and Snippet Modeling as key components.
Extensive experiments on four widely-used gait datasets validate the effectiveness of our proposed approach and, more importantly, highlight the potential of gait snippets.
For instance, our method achieves the rank-1 accuracy of 77.5\% on Gait3D and 81.7\% on GREW using a 2D convolution-based backbone.
\end{abstract} 

%% file: 1_introduction.tex

\section{Introduction}
\label{sec:intro}

Gait recognition aims to identify individuals based on their unique walking patterns.
This technique can be performed at a distance without the explicit cooperation of the subjects, making it highly applicable in areas such as social security~\cite{rida2019robust}, human-computer interaction~\cite{zhu20223d}, and health monitoring~\cite{bortone2021gait}, \etc.
Silhouettes are commonly used as input, as they effectively eliminate clothing texture while remaining robust under low-resolution conditions.

In the gait recognition literature, early studies typically aggregated silhouettes into a template, such as Gait Energy Image~\cite{han2005individual}, which, although simple, inevitably sacrifices fine-grained details.
Recent research predominantly treats silhouettes either as an unordered set or an ordered sequence, leveraging deep neural networks to extract gait features.
Specifically, set-based methods~\cite{chao2019gaitset,hou2020gait,hou2021setres,hou2022gait} assume that the appearance of a silhouette inherently contains its positional information, rendering the order information unnecessary.
The pioneering GaitSet~\cite{chao2019gaitset}, a representative of this category, significantly improves performance over template-based methods and demonstrates resilience to frame permutations.
In contrast, sequence-based methods~\cite{lin2020gait,lin2021gait,huang20213d,huang2021context} treat a sequence of silhouettes as a video, utilizing 3D~\cite{tran2015learning} or P3D~\cite{qiu2017learning} convolutions, along with their variants~\cite{lin2020gait}, to extract both spatial and temporal features.
%

Despite the significant performance gains of recent advancements, both set-based and sequence-based paradigms exhibit notable limitations.
First, in set-based methods, feature extraction in the backbone, typically performed using 2D convolution, processes each silhouette independently, lacking awareness of short-range temporal context between adjacent frames.
%
%
Second, in sequence-based methods, feature extraction primarily relies on 3D/P3D convolutions or their variants, with a limited number of continuous frames (\eg, $30$) sampled from each sequence during training.
This approach significantly hinders the ability to model long-range temporal dependencies, especially in long sequences (\eg, those with more than $200$ frames in real-world benchmarks~\cite{zheng2022gait}).
This raises a critical question: \emph{Is there an alternative paradigm for extracting gait features from silhouettes that addresses these limitations?}

\begin{figure}[t]
	\centering
	\includegraphics[width=0.9999\linewidth]{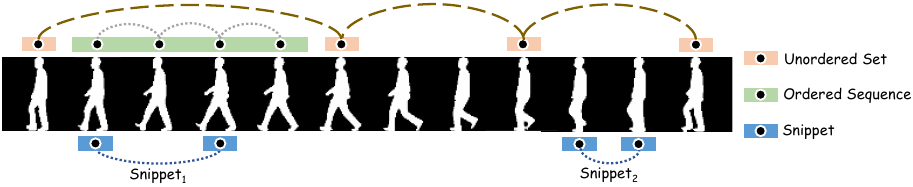}
	\caption{
		Illustration of gait snippets in comparison to unordered sets and ordered sequences.
		Best viewed in color.
	}
	\label{fig:snippet}
	\vspace{-2ex}
\end{figure}

%
%
In this work, we propose a new perspective on gait recognition inspired by human cognition, arguing that identification often depends on key actions in a few adjacent frames—not a full cycle.
This aligns with the biological finding that ``\emph{recognition is possible for stimuli lasting a fraction of a full walking cycle}''~\cite{giese2003neural}.
Motivated by this insight, we propose to conceptualize human gait as \emph{a composition of individualized actions}.
Specifically, as illustrated in Figure~\ref{fig:snippet}, we represent an action using several frames randomly selected from a continuous segment of the sequence, which we term a \textbf{snippet}.
This approach allows an individual’s walking pattern to be described as the union of snippets derived from the same sequence.
Gait snippets offer two notable conceptual advantages:
(1) Compared to unordered sets, snippets facilitate the incorporation of short-range temporal context for frame-level feature extraction.
(2) Compared to ordered sequences, snippets enable the capture of long-range temporal dependencies within a long sequence.

Building on these insights, we focus on snippet-based gait recognition and address two critical challenges:
(a) \emph{How to sample snippets during the input phase for training and inference?}
(b) \emph{How to effectively model snippet-based inputs for gait recognition?}
In this work, we propose an efficient yet effective solution, marking the first attempt to systematically tackle these challenges.

\emph{Regarding Snippet Sampling}, given a sequence of silhouettes, we treat it as non-continuous due to imperfect upstream processing and various occlusions~\cite{fan2022learning}, but we assume that the relative order of frames is preserved.
This order is used to divide the sequence into non-overlapping segments of equal length.
For training, we randomly select a subset of frames from each segment to form a snippet representing an individualized action, with the number of snippets generally fewer than the number of segments.
For inference, all frames from each segment are used to construct a snippet, and all snippets from a sequence are utilized to match the probe and gallery.
\emph{In terms of Snippet Modeling}, we design an efficient framework to address three core challenges:
(1) \textbf{Intra-Snippet Modeling}: We introduce a Snippet Block where a non-parametric pooling operation captures local temporal context within a snippet, merging it with frame-level features through a residual connection.
(2) \textbf{Cross-Snippet Modeling}: We treat all snippets within a sequence as an unordered set, employing Set Pooling to derive sequence-level representations based on intra-snippet modeling.
(3) \textbf{Snippet-Level Supervision}: Representing gait through snippets enables hierarchical representations at both the sequence and snippet levels.
In addition to sequence-level loss, we introduce snippet-level supervision to further enhance training.

In summary, the main contributions are threefold:
\begin{enumerate}[(1)]
  \item We introduce a new perspective on gait recognition, organizing a sequence of silhouettes as a union of snippets to characterize the walking pattern.
  \item We pioneer snippet-based gait recognition, designing a comprehensive solution that includes Snippet Sampling and Snippet Modeling.
  \item Extensive experimental results demonstrate the potential of gait snippets, with our approach achieving the rank-1 accuracy of 77.5\% on Gait3D~\cite{zheng2022gait} and 81.7\% on GREW~\cite{zhu2021gait} using a 2D convolutional backbone.
\end{enumerate}

%% file: 2_related_work.tex

\section{Related Work}
\label{sec:related}

\paragraph{Gait Recognition}
We address the fundamental challenges in the modeling paradigm for gait recognition by using silhouettes as input. 
In early studies~\cite{han2005individual,wang2010chrono}, silhouettes were usually aggregated into templates.
More recent advancements have treated silhouettes as either unordered sets~\cite{chao2019gaitset,hou2020gait,hou2021setres,hou2022gait,fan2023opengait} or ordered sequences~\cite{lin2020gait,fan2020gaitpart,lin2021gait,huang20213d,huang2021context,ma2023dynamic,dou2023gaitgci,wang2023hierarchical,wang2023dygait} for feature learning.
Below, we briefly review representative methods within these two subcategories.


(1) \emph{Unordered Sets}:
GaitSet~\cite{chao2019gaitset} is the first to introduce set-based feature learning for silhouettes, using horizontal splits of feature maps to learn multiple part representations.
GLN~\cite{hou2020gait} merges multi-stage features for set-based modeling, focusing on reducing feature dimensionality to enhance recognition performance.
GaitBase~\cite{fan2023opengait} and its deeper variant, DeepGaitV2-2D~\cite{fan2023exploring}, present a robust ResNet-like 2D backbone, achieving competitive performance across various benchmarks.

(2) \emph{Ordered Sequences}:
GaitGL~\cite{lin2021gait} utilizes 3D convolution to blend local and global feature extraction in its architecture.
GaitGCI~\cite{dou2023gaitgci} introduces a counterfactual intervention to mitigate the effects of confounding factors while using dynamic convolution for factual/counterfactual attention generation.
DyGait~\cite{wang2023dygait} captures dynamic features by leveraging differences between frame-level and template features.
DeepGaitV2-3D and DeepGaitV2-P3D~\cite{fan2023exploring} are variants of GaitBase~\cite{fan2023opengait} that utilize ordered input with 3D/P3D convolutions.
VPNet~\cite{ma2024learning} employs a ResNet50-like backbone for gait recognition and introduces visual prompts to handle complex variations in gait patterns.

%
%
%
%
%
%
%
%
%

\paragraph{Snippet Paradigm}
We noticed that the term “snippet” has been previously used in the action recognition literature~\cite{wang2016temporal,duan2023skeletr}, and we compare those approaches with our own.
For instance, TSN~\cite{wang2016temporal} constructs RGB snippets in a similar fashion but mandates that snippets be sampled from all segments and lacks intra-snippet modeling, which we consider crucial for snippet-based gait recognition.
SkeleTR~\cite{duan2023skeletr} processes short skeleton sequences as snippets but requires continuity within each snippet.
In our study, we extend the concept of snippets to gait recognition, where \emph{neither the frames within a snippet nor the snippets in a sequence need to be strictly continuous}.
Moreover, our approach diverges significantly from these methods by emphasizing snippet modeling, which will be elaborated in the next Section~\ref{sec:snippet_modeling}.

%% file: 3_approach.tex

\section{Our Approach}
\label{sec:approach}
In this work, we investigate a fundamental paradigm for gait recognition that addresses the limitations of unordered sets and ordered sequences.
Specifically, we propose a new perspective that treats human gait as \emph{a composition of individualized actions}, with each action represented by a \textbf{snippet}, which consists of a few frames randomly selected from a continuous segment of the sequence.
This snippet paradigm allows the model to leverage both short-range and long-range temporal contexts during training, enhancing its capability for comprehensive gait feature learning.

In the following sections, we will first describe our strategy for organizing a sequence of silhouettes into snippets.
Subsequently, we will present an effective approach to conduct snippet-based gait recognition.

\subsection{Snippet Sampling}
\label{sec:snippet_sampling}
The underlying principles of sampling strategies for gait recognition can generally be summarized from two perspectives:
(1) During training, a limited number of frames are typically sampled to represent a sequence due to the trade-off between computational cost and sampling diversity.
(2) During inference, all frames of a sequence are utilized to ensure accurate recognition.
Below, we briefly highlight the distinctions in sampling strategies when treating silhouettes as either unordered sets or ordered sequences.
Specifically, in the training phase, set-based methods randomly select \emph{discontinuous} frames from the entire sequence~\cite{chao2019gaitset}, whereas sequence-based methods select \emph{continuous} or \emph{nearly continuous} frames for temporal modeling~\cite{fan2020gaitpart}.

Our snippet-based sampling strategy influences both the training and inference phases, as described in detail in this section.
It is noteworthy that we assume \emph{the relative order of frames in a sequence is reliable, even though the frames themselves may not necessarily be continuous}, a condition that aligns well with practical applications~\cite{sepas2021deep,shen2022comprehensive}.

\begin{figure}[t]
	\centering
	\includegraphics[width=0.9999\linewidth]{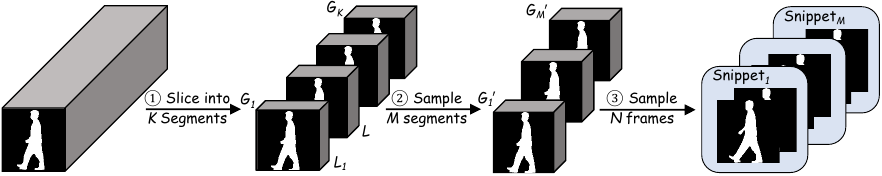}
	\caption{
		Snippet sampling for \emph{training}.
		$\{G_{1}, \cdots, G_{K}\}$ represent the total segments of a sequence, where $L$ is the segment length and $L_1$ for the first segment is a random integer to enhance sampling diversity.
		$\{G_{1}^{'}, \cdots, G_{M}^{'}\}$ represent the sampled segments.
		$M$ and $N$ denote the number of sampled snippets per sequence and the number of sampled frames per snippet, respectively.
	}
	\label{fig:snippet_sampling}
	\vspace{-2ex}
\end{figure}

\subsubsection{Training Phase}
During the training phase, we first partition a sequence into non-overlapping segments of equal duration, preserving the relative order, and then design the snippet sampling strategy based on three guiding principles:
(a) Given the constraints of computational resources and the need for sampling diversity, the total number of frames selected from a sequence should be limited, denoted as $S$.
(b) The fundamental unit within the sampled $S$ frames is a snippet, where each snippet consists of $N$ frames randomly selected from a segment to capture an individualized action.
(c) To increase sampling diversity and enhance model robustness, the segment partition for a sequence should vary across iterations.

Our approach is illustrated in Figure~\ref{fig:snippet_sampling}:
(1) A sequence of silhouettes is divided into $K$ segments, denoted as $\{G_{1}, G_{2}, \cdots, G_{K}\}$, each of length $L$, where $L$ typically approximates the number of frames in a gait cycle (\eg, $L\!=\!16$ in most cases~\cite{ma2024learning}).
If the sequence length is not perfectly divisible by $L$, the remaining frames are treated as an additional segment.
(2) When processing a sequence in a mini-batch, we randomly sample $M$ segments from it and then randomly select $N$ frames from each chosen segment to construct the snippets.
\emph{Sampling with replacement} is allowed when the number of segments or the number of frames in a segment is limited.
We ensure that $S \! = \! M \! \times \! N$, assigning each snippet a segment label $k$ ($k \! \in \! \{1, \cdots, K\}$) for subsequent modeling.
(3) To enhance sampling diversity within a sequence, the initial frames are treated as a special segment, with its length $L_{1}$ randomly chosen from $\{1, 2, \cdots, L\}$.

\subsubsection{Inference Phase}
The snippet sampling strategy for the inference phase is also developed based on three guiding principles:
(a) All frames in a sequence should be utilized to ensure precise matching between the probe and gallery.
(b) To maintain consistency with the training phase, sequences are divided into segments, with all frames in each segment forming a snippet.
(c) The segment partition should remain fixed to produce stable predictions.

Accordingly, our inference strategy involves the following three aspects:
(1) A sequence of silhouettes is divided into $K$ segments of equal length $L$, as previously defined in the training phase (\eg, $L \! = \! 16$).
(2) \emph{Each snippet comprises all frames within a segment}, and \emph{prediction features are extracted using all snippets from the sequence}, which is equivalent to setting $M \! = \! K$ and $N \! = \! L$ during inference.
(3) The length of the first segment $L_{1}$ is fixed to $L$, thereby eliminating the need for multiple forward passes and reducing inference overhead.

\begin{figure*}[t]
	\centering
	\includegraphics[width=0.9999\linewidth]{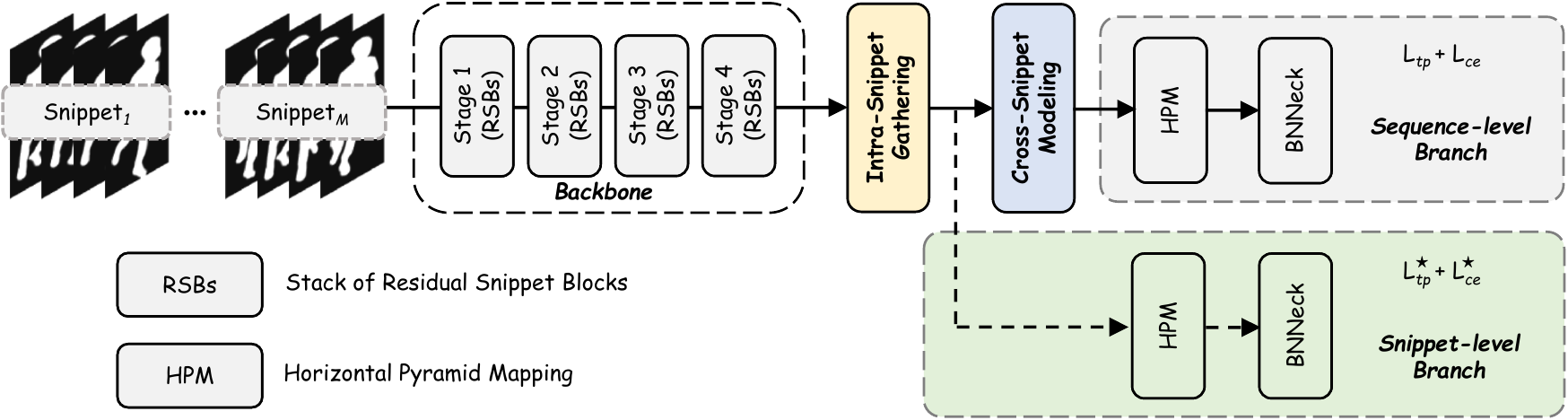}
	\caption{
		Illustration of GaitSnippet.
		(1) Residual Snippet Block \emph{integrating Intra-Snippet Modeling} as shown in Figure~\ref{fig:snippet_modeling}(b) serves as the basic component to construct the backbone.
		(2) At the end of the backbone, we first apply \emph{Intra-Snippet Gathering} (the \emph{Gathering} step for Intra-Snippet Modeling) to derive snippet-level representations and then perform \emph{Cross-Snippet Modeling} to obtain sequence-level representations.
		(3) In addition to sequence-level supervision, an auxiliary branch is introduced to enforce supervision on snippet-level features \emph{only for training}.
	}
	\label{fig:pipeline}
	\vspace{-2ex}
\end{figure*}

\subsection{Snippet Modeling}
\label{sec:snippet_modeling}
Snippets provide a new paradigm for modeling silhouettes in gait recognition. However, fully exploiting the potential advantages of snippets remains an open question. In this work, we propose an efficient yet effective solution to address this issue.
Specifically, we identify three primary challenges in snippet modeling for gait recognition: \textbf{Intra-Snippet Modeling}, \textbf{Cross-Snippet Modeling}, and \textbf{Snippet-Level Supervision}.
In the following sections, we systematically address these challenges through our proposed approach, which we term \textbf{GaitSnippet}.
The pipeline is illustrated in Figure~\ref{fig:pipeline}.

\subsubsection{Intra-Snippet Modeling}
\label{sec:intra_snippet}
\begin{figure}[t]
	\centering
	\includegraphics[width=0.55\linewidth]{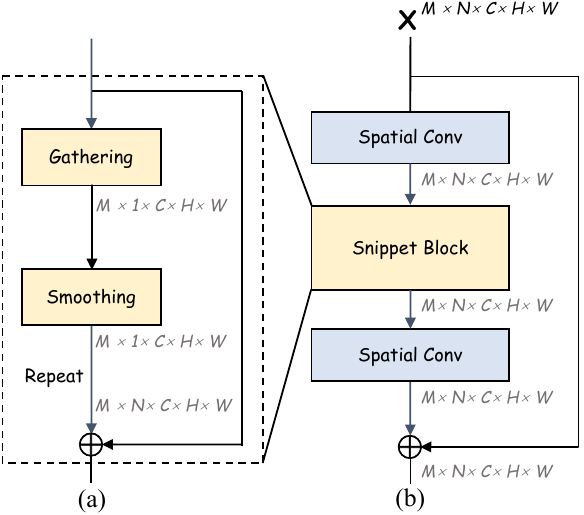}
	\caption{
		(a) Snippet Block.
		(b) Residual Snippet Block.
		$M$ and $N$ denote the number of snippets and the number of frames per snippet in a sequence, while $C$, $H$, and $W$ represent the dimensions of channel, height, and width.
	}
	\label{fig:snippet_modeling}
	\vspace{-2ex}
\end{figure}

In GaitSnippet, we address intra-snippet modeling with the objective of \emph{capturing local temporal context to enhance frame-level feature extraction} through a three-step process:
\begin{enumerate}[(1)]
  \item \emph{Gathering:} Considering that the frames within a snippet are not necessarily continuous, we treat a snippet as an unordered set.
  Based on this formulation, we utilize the efficient Set Pooling technique to aggregate the features of a snippet, which is implemented through a non-parametric Temporal Max Pooling operation~\cite{chao2019gaitset}.
  \item \emph{Smoothing:} To mitigate the negative effects of noise within a snippet and reduce the semantic gap between different levels of features, we apply a smoothing layer, typically implemented using a $1 \times 1$ convolution, following the \emph{Gathering} step.
  \item \emph{Residual:} To make frame-level feature extraction aware of local temporal context in a snippet, we incorporate a residual connection to merge the snippet-level output after smoothing with the frame-level features of the corresponding snippets.
\end{enumerate}
As illustrated in Figure~\ref{fig:snippet_modeling}(a), these steps are formulated into a basic block called \textbf{Snippet Block}.

Furthermore, recent advancements in gait recognition have demonstrated that a plain 2D residual backbone~\cite{fan2023opengait,fan2023exploring} can achieve highly competitive performance in both constrained and unconstrained environments, while maintaining significantly lower computational costs compared to their 3D counterparts.
The spatial convolution, specifically applied along the height and width dimensions, plays a critical role in extracting frame-level features.
To facilitate effective collaboration between intra-snippet modeling and spatial convolution, we draw inspiration from P3D~\cite{qiu2017learning} and integrate a Snippet Block between two spatial convolutional layers within a standard residual block.
The rationale behind this approach is to \emph{enable each frame to become aware of local temporal context within a snippet during successive stages of frame-level feature extraction}.
Ultimately, the architecture illustrated in Figure~\ref{fig:snippet_modeling}(b), called \textbf{Residual Snippet Block}, serves as the basic component to construct the backbone for GaitSnippet as shown in Figure~\ref{fig:pipeline}.

\subsubsection{Cross-Snippet Modeling}
\label{sec:cross_snippet}
For cross-snippet modeling, our objective is to \emph{acquire a robust global representation for a gait sequence based on the snippet-level features}.
As a pioneering attempt and to ensure a fair comparison with the base models~\cite{fan2023opengait,fan2023exploring}, we conduct cross-snippet modeling on the output of the backbone which corresponds to the \emph{frame-level} features.
Specifically, we first apply \emph{Intra-Snippet Gathering} (the \emph{Gathering} step for intra-snippet modeling) on the frame-level features to derive snippet-level representations.
Subsequently, we treat all snippets from a sequence as an unordered set and employ another Set Pooling~\cite{chao2019gaitset} to perform cross-snippet modeling.
In practice, this is implemented using Temporal Max Pooling on all \emph{snippet-level} representations within a sequence.
%

It is crucial to highlight that (1) GaitSnippet involves two \emph{hierarchical} unordered sets: frames within a snippet and all snippets within a sequence.
However, the snippet-based modeling approach is \emph{not} permutation-invariant to the frame order, distinguishing it from methods that exclusively rely on unordered sets~\cite{chao2019gaitset,fan2023opengait}.
Unlike unordered sets, the use of snippets enables the exploitation of local temporal context in frame-level feature extraction, which is vital for learning discriminative and complementary features for individual silhouettes.
(2) At the end of the backbone, Temporal Max Pooling employed for both intra-snippet and cross-snippet modeling makes the sequence-level features equivalent to the maximum of all frames.
%
Yet the intermediate output from intra-snippet modeling is essential for enabling Snippet-Level Supervision.
Further discussion about the role of Temporal Max Pooling is provided in Section~\ref{sec:discuss_modeling} of the appendix.

\subsubsection{Snippet-Level Supervision}

The snippet-based modeling of gait conveniently facilitates the extraction of two hierarchical representations for a sequence, namely, \emph{sequence-level} and \emph{snippet-level} representations.
For supervision on the \emph{sequence-level representations}, we adopt the typical approach outlined in~\cite{fan2023opengait}.
Initially, Horizontal Pyramid Mapping~\cite{fu2019horizontal,chao2019gaitset} (including linear layers for separate parts) is utilized to horizontally split the features for obtaining fine-grained part representations efficiently.
Then, for each part, we employ triplet loss $\mathcal{L}_{tp}$ and cross-entropy loss $\mathcal{L}_{ce}$, assisted by BNNeck~\cite{luo2019bag}, for training.
Formally, these losses are defined as follows:
\begin{equation}
\label{eq_loss_sequence}
\small
\begin{aligned}
\mathcal{L}_{tp} \! = \!
\frac{1}{N_{tp}} &
\overbrace{ \sum_{u=1}^{U} \sum_{v=1}^{V} }^{anchor}
\overbrace{ \sum_{a=1}^{V} }^{pos}
\overbrace{ \sum_{ \substack{b=1\\b \ne u} }^{U}  \sum_{c=1}^{V} }^{neg}
\! \left[
\delta \! + \! \mathcal{D}(\mathcal{F}_{u,v}, \mathcal{F}_{u,a}) \! - \! \mathcal{D}(\mathcal{F}_{u,v}, \mathcal{F}_{b,c})
\right]_{+}
\\
& \mathcal{L}_{ce} \! = \!
- \frac{1}{U \times V}
\overbrace{ \sum_{u=1}^{U}  \sum_{v=1}^{V}}^{batch}
\overbrace{ \sum_{c=1}^{N_{c}}}^{sub}
q_{u,v,c} \log p_{u,v,c}
\end{aligned}
\end{equation}
Here, \emph{pos}, \emph{neg}, and \emph{sub} stand for \emph{positive}, \emph{negative}, and \emph{subjects}, respectively.
$(U, V)$ denote the number of subjects and the number of sequences per subject in a mini-batch.
$N_{tp}$ serves as a normalization coefficient accounting for the non-zero triplet terms.
$\delta$ is a margin threshold and $[~]_{+}$ works as the ReLU function.
$\mathcal{F}$ denotes the sequence-level representations and $\mathcal{D}$ measures the Euclidean distance.
$(\mathcal{F}_{u,v}, \mathcal{F}_{u,a})$ and $(\mathcal{F}_{u,v}, \mathcal{F}_{b,c})$ represent positive and negative pairs, respectively.
$N_{c}$ is the number of subjects in the training set, while $p$ and $q$ denote the predicted probabilities and the one-hot ground-truth identity labels.

%
With the snippet-based approach to gait, we can conveniently obtain snippet-level representations in addition to sequence-level representations, motivating us to introduce the following fine-grained supervision.
Specifically, we add a separate branch to process snippet-level representations prior to cross-snippet modeling, using Horizontal Pyramid Mapping to obtain part-level features and incorporating BNNeck analogous to the sequence-level branch.
Formally, for each part, the snippet-level triplet loss $\mathcal{L}_{tp}^{\star}$ and cross-entropy loss $\mathcal{L}_{ce}^{\star}$ are computed as follows:
\begin{equation}
\label{eq_loss_snippet}
\small
\begin{aligned}
\mathcal{L}_{tp}^{\star} \! = \!
\frac{1}{N_{tp}^{\star}}
\overbrace{ \overbrace{\sum_{u=1}^{U}  \sum_{v=1}^{V}}^{batch}  \overbrace{\sum_{m=1}^{M}}^{snp} }^{anchor} &
\overbrace{ \sum_{a=1}^{V}  \overbrace{\sum_{i=1}^{M}}^{snp} }^{pos}
\overbrace{ \sum_{ \substack{b=1\\b \ne u} }^{U}  \sum_{c=1}^{V}  \overbrace{\sum_{j=1}^{M}}^{snp} }^{neg}
\! \left[
\delta \! + \! \mathcal{D}(\mathcal{F}^{\star}_{u,v,m}, \mathcal{F}^{\star}_{u,a,i}) \! - \! \mathcal{D}(\mathcal{F}^{\star}_{u,v,m}, \mathcal{F}^{\star}_{b,c,j})
\right]_{+}
\\
\mathcal{L}_{ce}^{\star} \! = \!
- & \frac{1}{U \! \times \! V \! \times \! M}
\overbrace{ \sum_{u=1}^{U}  \sum_{v=1}^{V}}^{batch}
\overbrace{ \sum_{m=1}^{M}}^{snp}
\overbrace{ \sum_{c=1}^{N_{c}}}^{sub}
q^{\star}_{u,v,m,c} \log p^{\star}_{u,v,m,c}
\end{aligned}
\end{equation}
where \emph{snp} denotes snippets, $M$ is the number of sampled snippets per sequence for training, and $\mathcal{F}^{\star}$ refers to snippet-level representations.
The remaining symbols are similar to those in Eq.~\ref{eq_loss_sequence}, with the superscript $\star$ indicating snippet-level computations.

We then define the integrated objective for one of the part representations as:
\begin{equation}
\label{eq_loss_all}
\small
\begin{aligned}
\mathcal{L}_{all} &= \mathcal{L}_{tp} + \mathcal{L}_{ce} + \alpha \times (\mathcal{L}^{\star}_{tp} + \mathcal{L}^{\star}_{ce})
\end{aligned}
\end{equation}
where $\alpha$ is a hyperparameter to balance the two levels of supervision signals.
%
The final loss is computed by averaging the above losses across all parts, which is used to train the entire network.

It is important to emphasize that the additional branch for snippet-level supervision is employed \emph{only during the training phase}, thereby leaving the inference complexity unaffected.
For evaluation, we utilize the features extracted before BNNeck in the sequence-level branch to compute similarities between the probe and gallery sequences.

%% file: 4_experiments.tex
\section{Experiments}
\label{sec:experiments}

\subsection{Settings}
We conduct experiments on four widely-used gait datasets: Gait3D~\cite{zheng2022gait} and GREW~\cite{zhu2021gait}, CCPG~\cite{li2023depth} and CCGR-MINI~\cite{zou2024cross}.
In the training phase, we adopt $L \! = \! 16$ to approximate the number of frames depicting a gait cycle~\cite{ma2024learning} for segment partition, and $L_1$ is a random integer sampled from $\{1, 2, \cdots, 16\}$.
To sample a sequence, we randomly select $M \! = \! 4$ snippets and $N \! = \! 8$ frames per snippet, \ie, we sample $S \! = \! 32$ frames for each sequence.
For evaluation, we set $L_1 \! = \! L \! = \! 16$ for segment partition.
All frames in a segment are treated as a snippet, and all snippets for a sequence are used to extract gait features.
Detailed dataset statistics and implementation details are provided in the appendix.

\begin{table}[!tbp]
	\setlength{\tabcolsep}{8pt}
	\begin{center}
		\resizebox{0.8\linewidth}{!}{%
			\begin{tabular}{c|c|c|cc|cc}
				\hline
				\multirow{2}{*}{Method} &  \multirow{2}{*}{\tabincell{c}{Cate-\\gory}} &  \multirow{2}{*}{\tabincell{c}{Back-\\bone}} &  \multicolumn{2}{c|}{Gait3D} &  \multicolumn{2}{c}{GREW} \\
				&	&   & R1   & mAP   & R1   & R5     \\
				\hline
				GaitPart~\cite{fan2020gaitpart} & \multirow{11}{*}{\emph{Seq}} & 2D & 28.2 & 21.6 & 47.6 & 60.7 \\
				GaitGL~\cite{lin2021gait} & & 3D & 29.7 & 22.3 & 47.3 & 63.6 \\
				GaitGCI~\cite{dou2023gaitgci} & & 3D & 50.3 & 39.5 & 68.5 & 80.8 \\
				DyGait~\cite{wang2023dygait} & & 3D & 66.3 & 56.4 & 71.4 & 83.2 \\
				HSTL~\cite{wang2023hierarchical} & & 3D & 61.3 & 55.5 & 62.7 & 76.6 \\
				SwinGait-3D~\cite{fan2023exploring} & & Swin3D & 75.0 & \RED{67.2} & 79.3 & 88.9 \\
				DeepGaitV2-3D~\cite{fan2023exploring} & & 3D & 72.8 & 63.9 & 79.4 & 88.9 \\
				DeepGaitV2-P3D~\cite{fan2023exploring} & & P3D & 74.4 & 65.8 & 77.7 & 87.9 \\
				VPNet~\cite{ma2024learning} & & 3D & \RED{75.4} & / & \RED{80.0} & \RED{89.4} \\
				CLTD~\cite{xiongcausality} & & 3D & 69.7 & / & 78.0 & 87.8 \\
				GaitMoE~\cite{huang2025occluded} & & 3D & 73.7 & 66.2 & 79.6 & 89.1 \\
				\hline
				GaitSet~\cite{chao2019gaitset} & \multirow{4}{*}{\emph{Set}} & 2D & 36.7 & 30.0 & 48.4 & 63.6 \\
				GaitBase~\cite{fan2023opengait} & & 2D & 64.6 & 55.3 & 60.1 & 75.5 \\
				SwinGait-2D~\cite{fan2023exploring} & & Swin2D & \BLUE{69.4} & \BLUE{61.6} & \BLUE{70.8} & \BLUE{83.7} \\
				DeepGaitV2-2D~\cite{fan2023exploring} & & 2D & 68.2 & 60.4 & 68.6 & 82.0 \\
				\hline
				GaitSnippet~(\bftab{Ours})  & \bftab{\emph{Snippet}} & 2D & \bftab{77.5} & \bftab{69.4} & \bftab{81.7} & \bftab{90.9} \\
				\hline
			\end{tabular}
		}
	\end{center}
	\caption{
		Performance comparison on Gait3D~\cite{zheng2022gait} and GREW~\cite{zhu2021gait}.
		The results are reported in rank-1 (R1, \%), rank-5 (R5, \%), and mean Average Precision (mAP, \%).
		The best results in each category are marked in \RED{red}, \BLUE{blue}, and \bftab{bold}, respectively.
	}
	\label{tab:comparison_wild}
\vspace{-2ex}
\end{table}

\subsection{Performance Comparison}
\paragraph{Gait3D \& GREW}
The emergence of Gait3D and GREW has advanced gait recognition research from controlled laboratory settings to real-world environments.
%
%
In Table~\ref{tab:comparison_wild}, we present a performance comparison on in-the-wild benchmarks. The methods are categorized into three groups based on how they treat the input: \emph{Ordered Sequences}, \emph{Unordered Sets}, and the brand-new \emph{Snippets}.

From the results in Table~\ref{tab:comparison_wild}, the following observations can be made:
(1) Sequence-based methods achieve state-of-the-art performance and mostly employ 3D or P3D convolution in the backbone, which generally entails higher computational costs compared to 2D convolution-based backbones.
%
%
(2) DeepGaitV2-2D~\cite{fan2023exploring}, despite their simplicity, achieve highly competitive performance on these benchmarks.
%
%
(3) GaitSnippet outperforms advanced methods on both benchmarks using a 2D convolutional backbone.
%
Specifically, the performance gains compared to 2D convolution-based baselines (\eg, R1: +9.3\%, mAP: +9.0\% over DeepGaitV2-2D on Gait3D with the same network depth) effectively demonstrate the effectiveness of snippet-based gait recognition.

\paragraph{CCPG \& CCGR-MINI}
With the increasing interest in gait recognition, several new datasets have recently been introduced~\cite{li2023depth,shen2023lidargait,li2024aerialgait,zou2024cross}, aiming to address more diverse and challenging scenarios.
To further demonstrate the generalizability of GaitSnippet, we additionally evaluate it on two representative emerging datasets: CCPG~\cite{li2023depth}\footnote{We re-ran the experiments on CCPG using the same preprocessing and protocols as the baselines~\cite{fan2023exploring,fan2023opengait} to ensure fair comparisons. The results differ slightly from those in the initial submission, but they remain state of the art.} and CCGR-MINI~\cite{zou2024cross}.
As shown in Table~\ref{tab:comparison_emerging}, GaitSnippet achieves state-of-the-art performance on both datasets, further validating the effectiveness and adaptability of snippet-based modeling for gait recognition.

\subsection{Ablation Study}
%

\subsubsection{Ablation Study on Snippet Sampling}
%
In Table~\ref{tab:ablation_sampling}, we present an ablation study on Snippet Sampling from two perspectives.

First, we evaluate the overall effect of the sampling strategy using our base model (\ie, DeepGaitV2-2D) in the first part, including set-based, sequence-based, and snippet-based strategies.
Interestingly, from the first three rows, we observe that Snippet Sampling during training also benefits recognition performance with DeepGaitV2-2D, which is based on unordered sets.
A likely reason for this is that our sampling strategy enhances the robustness of DeepGaitV2-2D by narrowing the distribution gap between discontinuous frames during training and continuous sequences during testing.

Second, we analyze the effect of hyper-parameters for Snippet Sampling in the second part.
Specifically, during the training phase, there are four hyper-parameters for sampling snippets from a sequence: $L$--the segment length, $S$--the total number of frames sampled from a sequence, $M$--the number of snippets sampled from a sequence, and $N$--the number of frames sampled per snippet from a segment.
Note that $S \! = \! M \! \times \! N$ is always maintained.
During our experiments with Snippet Sampling, we fix $S \! = \! 32$, considering computational cost and ensuring fair comparisons. 
In Table~\ref{tab:ablation_sampling}, we conduct ablation studies on the other sampling parameters.
(1) We set $L \! = \! 16$ to approximate the number of frames in a gait cycle~\cite{ma2024learning}, and we also tried $L \! \in \! \{8, 32\}$ in the fourth/fifth rows.
(2) We set $N \! = \! 8$, which is half of a gait cycle, and also tried $N \! \in \! \{4, 16\}$ in the sixth/seventh rows.
(3) Given that $S \! = \! M \! \times \! N$, $M$ varies with different values of $N$ in each case (\ie, $M \! = \! 4$ for $N \! = \! 8$, $M \! = \! 8$ for $N \! = \! 4$, and $M \! = \! 2$ for $N \! = \! 16$).

\begin{table}[!tbp]
\setlength{\tabcolsep}{8pt}
	\begin{center}
		\resizebox{0.99\linewidth}{!}{%
			\begin{tabular}{c|c|c|ccccc|cc}
			\hline
			\multirow{2}{*}{Method} &  \multirow{2}{*}{\tabincell{c}{Category}} &  \multirow{2}{*}{\tabincell{c}{Backbone}} &  \multicolumn{5}{c|}{CCPG} &  \multicolumn{2}{c}{CCGR-MINI} \\
    		&	& & CL & UP & DN & BG & AVG & R1   & mAP   \\
			\hline
GaitPart~\cite{fan2020gaitpart} & \multirow{2}{*}{\emph{Seq}} & 2D & 79.2 & 85.3 & 86.5 & 88.0 & 84.8 & 8.0 & 10.1 \\
DeepGaitV2-P3D~\cite{fan2023exploring} & & P3D & \RED{90.5} & \RED{96.3} & \RED{91.4} & \RED{96.7} & \RED{93.7} & \RED{39.4} & \RED{36.0} \\
			\hline
GaitSet~\cite{chao2019gaitset} & \multirow{2}{*}{\emph{Set}} & 2D & 77.5 & 85.0 & 82.9 & 87.5 & 83.2 & 13.8 & 15.4 \\
GaitBase~\cite{fan2023opengait} & & 2D & \BLUE{88.5} & \BLUE{92.7} & \BLUE{93.4} & \BLUE{93.2} & \BLUE{92.0} & \BLUE{27.0} & \BLUE{24.9} \\
			\hline
GaitSnippet~(\bftab{Ours})  & \bftab{\emph{Snippet}} & 2D & \bftab{91.5} & \bftab{96.6} & \bftab{94.6} & \bftab{97.7} & \bftab{95.1} & \bftab{42.4} & \bftab{39.5} \\
			\hline
			\end{tabular}
		}
	\end{center}
    \caption{
    	Performance comparison on CCPG~\cite{li2023depth} and CCGR-MINI~\cite{zou2024cross}.
    	The results on CCPG are reported in rank-1 (R1, \%) accuracy, while those on CCGR-MINI are reported in rank-1 accuracy (R1, \%) and mean Average Precision (mAP, \%).
    	CL, UP, DN, BG, and AVG refer to changing full outfits, changing top clothes, changing pants, walking with bags, and mean accuracy, respectively.
    	The best performance in each category is highlighted in \RED{red}, \BLUE{blue}, and \bftab{bold}, respectively.
    }
    \label{tab:comparison_emerging}
\vspace{-2ex}
\end{table}

\subsubsection{Ablation Study on Snippet Block}
%
In this section, we analyze the three steps involved in intra-snippet modeling, namely \emph{Gathering}, \emph{Smoothing}, and \emph{Residual}. 
The results are presented in the first part of Table~\ref{tab:ablation_modeling}. 

To clarify the results:
(1) If none of the techniques in Table~\ref{tab:ablation_modeling} is applied, GaitSnippet is equivalent to DeepGaitV2-2D with Snippet Sampling.
(2) In the first row, when \emph{Gathering} is removed for each stage of intra-snippet modeling, snippet-level supervision becomes inapplicable.
(3) In the second row, the experiment highlights the importance of the smoothing layer, which acts as a bridge between frame-level features and those aggregated from a snippet.
When \emph{Smoothing} is removed, \emph{the model for inference does not introduce any additional parameters}, as the snippet-level branch is only employed during training.
In this case, the local context modeling within each Snippet Block still works but is more susceptible to the noise within a snippet and the semantic gap between different levels of features.
(4) For the third row, it is worth emphasizing that \emph{Residual} in Table~\ref{tab:ablation_modeling} refers to \emph{the integration of local contextual information from a snippet with frame-level features as depicted in Figure~\ref{fig:snippet_modeling}(a)}, rather than the standard residual connection shown in Figure~\ref{fig:snippet_modeling}(b).
When \emph{Residual} is removed, only the contextual information from the snippet is used for subsequent layers, which inevitably results in the loss of fine-grained details from each silhouette after the first Snippet Block.
%
%

\subsubsection{Ablation Study on Snippet-Level Supervision}
Benefitting from the snippet paradigm, we can easily incorporate snippet-level losses to provide fine-grained supervision for gait feature learning.
In the second part of Table~\ref{tab:ablation_modeling}, we perform ablation studies to analyze the effect of snippet-level supervision.
We can observe that:
(1) When $\alpha=0$, as shown in the fifth row, only the sequence-level loss is used to train the entire model, and our approach still achieves highly competitive performance.
(2) Snippet-level supervision improves recognition performance with varied loss weights.

Additionally, in the last row of Table~\ref{tab:ablation_modeling}, we experiment with sharing weights between the sequence-level and snippet-level branches shown in Figure~\ref{fig:pipeline}, consisting of Horizontal Pyramid Mapping and BNNeck.
The corresponding results show a moderate performance degradation, likely due to the semantic gap between the sequence-level and snippet-level features, making weight sharing between the branches inappropriate.
%

\begin{table}[!tbp]
	\setlength{\tabcolsep}{8pt}
	\begin{center}
		\resizebox{0.75\linewidth}{!}{%
			\begin{tabular}{c|c|c|c|c|c|c}
				\hline
				Model & \tabincell{c}{Sampling\\Strategy} & ~~~$L$~~~ & ~~~$M$~~~ & ~~~$N$~~~ & R1 & mAP \\
				\hline
				\hline
				\multirow{3}{*}{DeepGaitV2-2D} 		& \emph{Set} 		& -  & - & - & 68.2 	& 60.4  \\
				& \emph{Seq} 		& -  & - & - & 66.0 	& 58.7  \\
				& \emph{Snippet}    & 16 & 4 & 8 & 69.5 	& 61.5  \\
				\hline
				\hline
				\multirow{5}{*}{GaitSnippet} & \multirow{5}{*}{\emph{Snippet}}
				& 8 & 4 & 8  				& 76.4	& 69.0 \\
				& & 32 & 4 & 8  			& 74.7 	& 67.5 \\
				\cline{3-7}
				& & 16 & 8 & 4  			& 74.3	& 66.7 \\
				& & 16 & 2 & 16  			& 75.2 	& 66.9 \\
				\cline{3-7}
				& & \bftab{16} & \bftab{4} & \bftab{8}  & \bftab{77.5} 	& \bftab{69.4} \\
				\hline
			\end{tabular}
		}
	\end{center}
	\caption{
		Ablation study on snippet sampling.
		$L$, $M$, and $N$ denote the segment length for sequence partition, the number of snippets sampled per sequence, and the number of frames sampled per snippet, respectively.
		%
		%
		The results are reported on Gait3D.
	}
	\label{tab:ablation_sampling}
\end{table}

\begin{table}[!tbp]
	\setlength{\tabcolsep}{8pt}
	\begin{center}
		\resizebox{0.75\linewidth}{!}{%
			\begin{tabular}{c|c|c|c|c|c|c}
				\hline
				\multicolumn{3}{c|}{Snippet Block} & \multicolumn{2}{c|}{Snippet Supervision} & \multirow{2}{*}{R1} & \multirow{2}{*}{mAP} \\
				\cline{1-5}
				\emph{Gathering} & \emph{Smoothing} & \emph{Residual} & ~~~~~~\emph{$\alpha$}~~~~~~  & \emph{SW} & & \\
				\hline
				$\times$ 	& $\surd$ 	& $\surd$ 	& - & - & 73.3    & 65.7 \\
				$\surd$ 	& $\times$ 	& $\surd$ 	& 0.75 & $\times$ & 74.8    & 66.6 \\
				$\surd$ 	& $\surd$ 	& $\times$ 	& 0.75 & $\times$ & 72.5    & 63.7 \\
				$\bm{\surd}$ 	& $\bm{\surd}$ 	& $\bm{\surd}$ 	& \bftab{0.75} & $\bm{\times}$ & \bftab{77.5}  & \bftab{69.4} \\
				\hline
				\hline
				$\surd$ 	& $\surd$ 	& $\surd$ 	& 0.00 & $\times$ & 75.8    & 68.5 \\
				$\surd$ 	& $\surd$ 	& $\surd$ 	& 0.50 & $\times$ & 76.4    & 69.4 \\
				$\surd$ 	& $\surd$ 	& $\surd$ 	& 1.00 & $\times$ & 76.6    & 69.4 \\
				$\surd$ 	& $\surd$ 	& $\surd$ 	& 0.75 & $\surd$ & 75.5    & 68.8 \\
				\hline
			\end{tabular}
		}
	\end{center}
	\caption{
		Ablation study on snippet modeling on Gait3D.
		\emph{Gathering}, \emph{Smoothing}, and \emph{Residual} are the three steps in a Snippet Block.
		$\alpha$ represents the loss weight for snippet-level losses, and $SW$ denotes sharing weights between the sequence-level and snippet-level branches in Figure~\ref{fig:pipeline}.
		The experiments are conducted on Gait3D.
	}
	\label{tab:ablation_modeling}
\vspace{-2ex}
\end{table}

%% file: 5_conclusion.tex

\section{Conclusion}
\label{sec:conclusion}
In this work, we explore a new paradigm for gait recognition that integrates the strengths of unordered sets and ordered sequences.
Motivated by the observation that human identification does not necessarily rely on a complete gait cycle~\cite{giese2003neural}, we conceptualize human gait as a combination of individualized actions, with each action represented by a few frames that are adjacent but not necessarily continuous.
In essence, gait snippets enable the model to simultaneously exploit both short-range and long-range temporal contexts, which is beneficial for learning identity-related features from entire walking sequences.
Furthermore, we provide a non-trivial solution based on gait snippets, addressing the challenges of Snippet Sampling and Snippet Modeling.
Extensive experiments across various benchmarks demonstrate that our approach consistently improves performance and achieves state-of-the-art results in the wild, effectively verifying the potential of snippet-based gait recognition.

\section*{Ethical Statement}
The datasets used in our experiments are widely adopted in the literature, with informed consent obtained from all subjects during data collection. Additionally, no personal identifiers are accessible.
We strongly advocate that research in this field should be conducted with strict privacy protection measures in place.

\section*{Reproducibility Statement}
All experiments are conducted on publicly available gait recognition datasets (Gait3D, GREW, CCPG and CCGR-MINI) following the official splits and evaluation protocols.
Detailed hyperparameters and training procedures are provided in the paper.
The complete code and pre-trained models will be released.

\section*{Acknowledgement}
\label{sec:acknowledgement}
This work is jointly supported by National Natural Science Foundation of China (62276025, 62476027) and the Fundamental Research Funds for the Central Universities (2253200026).

%% file: X_supp.tex
\newpage

\section{Appendix}
\label{sec:appendix}

\subsection{Dataset Details}
The statistics for the widely-used datasets employed in our research, namely Gait3D~\cite{zheng2022gait}, GREW~\cite{zhu2021gait}, CCPG~\cite{li2023depth} and CCGR-MINI~\cite{zou2024cross}, are presented in Table~\ref{tab:datasets}.

\textbf{Gait3D} is a large-scale benchmark dataset captured in a supermarket environment, with two two-hour video segments randomly selected from each of seven days.
During evaluation, one sequence per subject is designated as the probe, while the remaining sequences are utilized as the gallery.

\textbf{GREW} is collected from multiple cameras in an uncontrolled environment over the course of a single day, resulting in diverse view variations.
Following the official evaluation protocol, each subject has four sequences, with two sequences used as the probe and the remaining two as the gallery.

\textbf{CCPG} is a cloth-changing benchmark dataset for person re-identification and gait recognition. It includes sequences of subjects captured in indoor and outdoor scenes, with the subjects having different clothing variations.
Following the standard protocol, the subjects are divided into two parts: the first half is used for training, and the remaining data is used for testing.

\textbf{CCGR-MINI} is a subset of the Cross-Covariate Gait Recognition (CCGR) dataset, specifically designed to address covariate diversity at both population and individual levels.
CCGR-MINI retains the diversity while enabling efficient evaluation and training under limited computational budgets.

\begin{table}[!tbp]
	\begin{center}
		\resizebox{0.75\linewidth}{!}{%
			\begin{tabular}{c|cc|cc|c|c}
				\hline
				\multirow{2}{*}{Dataset} & \multicolumn{2}{c|}{Train Set} &  \multicolumn{2}{c|}{Test Set} &  \multirow{2}{*}{\tabincell{c}{Walking\\Condition}} &  \multirow{2}{*}{\#Cam} \\
				& \#ID & \#Seq  & \#ID & \#Seq  &  & \\
				\hline
				Gait3D    & 3000                       & 18940  & 1000    & 6369   & Diverse               & 39  \\
				GREW      & 20000                      & 102887 & 6000    & 24000  & Diverse               & 882 \\
                CCPG      & 100                        & 8187   & 100     & 8379   & NM/BG/CL 			   & 10 \\
				CCGR-MINI & 570                        & 27507  & 400     & 20377  & Diverse               & 33  \\
				\hline
			\end{tabular}
		}
	\end{center}
	\caption{
		Dataset statistics.
		For each dataset, we present the number of subjects (\#ID) and sequences (\#Seq), walking conditions (NM/BG/CL for normal walking, walking with bags, and walking in different clothes), and the number of cameras (\#Cam).
	}
	\label{tab:datasets}
\end{table}
\begin{table}[!tbp]
	\setlength{\tabcolsep}{12pt}
	\begin{center}
		\resizebox{0.8\linewidth}{!}{%
			\begin{tabular}{c|c|c|c}
				\hline
				Dataset & \emph{blocks} & \emph{channels}  & \emph{strides} \\
				\hline
				Gait3D & [1, 4, 4, 1] & [64, 128, 256, 512] & [1, 2, 2, 1] \\
				GREW & [3, 4, 6, 3] & [64, 128, 256, 512] & [1, 2, 2, 1] \\
				CCPG & [1, 1, 1, 1] & [64, 128, 256, 512] & [1, 2, 2, 1] \\
                CCGR-MINI & [1, 4, 4, 1] & [64, 128, 256, 512] & [1, 2, 2, 1] \\
				\hline
			\end{tabular}
		}
	\end{center}
	\caption{
		The backbone settings for each dataset.
		\emph{Blocks}, \emph{channels}, and \emph{strides} refer to the number of blocks, convolutional channels, and strides for all stages, respectively.
		We configure the snippet-based backbone for each dataset with reference to the sequence-based counterparts~\cite{fan2023exploring,ma2024learning}.
	}
	\label{tab:settings}
\vspace{-2ex}
\end{table}

\begin{figure}[t]
	\centering
	\includegraphics[width=0.75\linewidth]{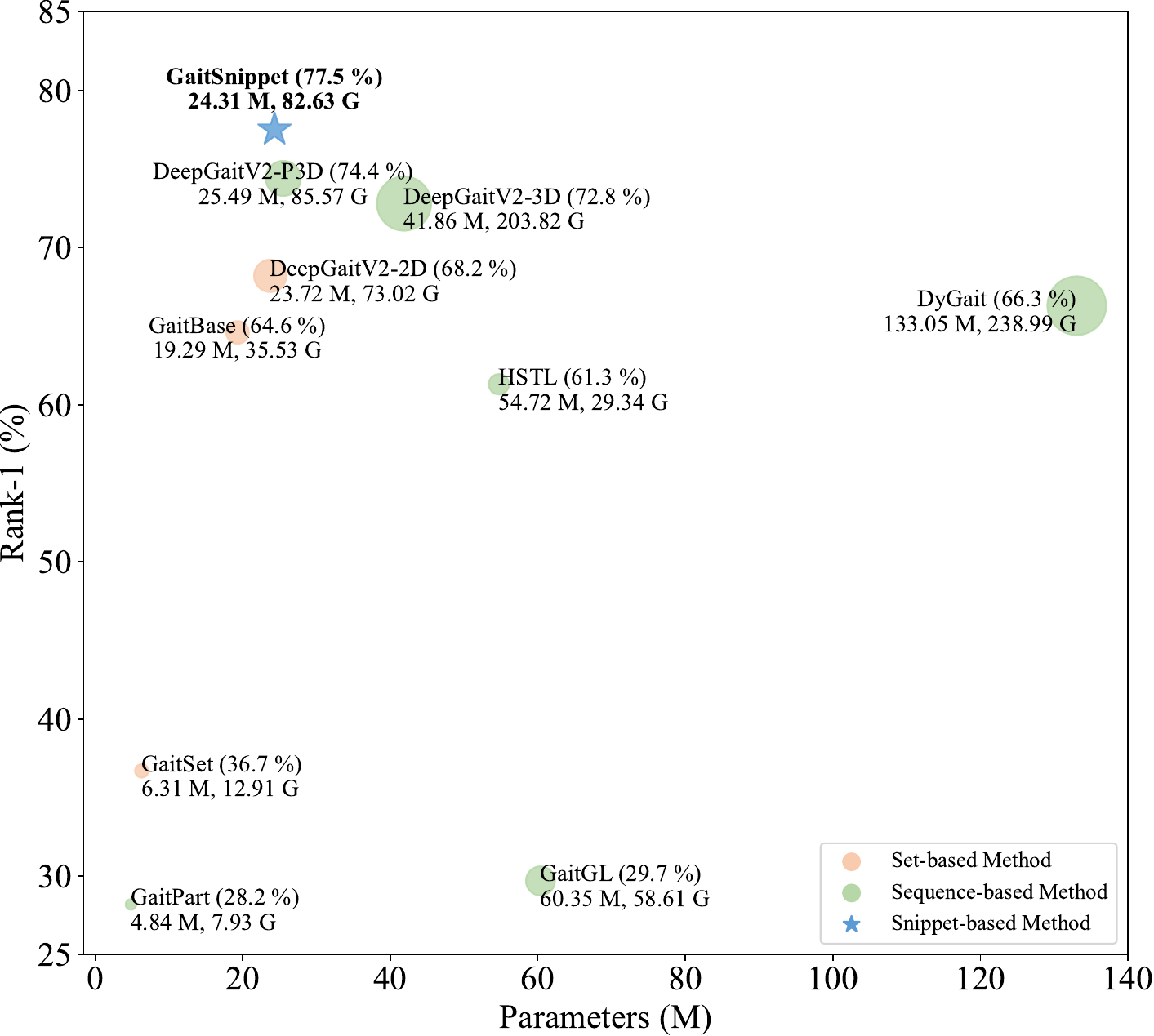}
	\caption{
		Computation cost in terms of parameters and FLOPs.
		The statistics are obtained on Gait3D, following the methodology of~\cite{wang2023hih,huang2025occluded}.
	}
	\label{fig:ablation_cost}
\vspace{-2ex}
\end{figure}

\subsection{Implementation Details}
Residual Snippet Block in Figure~\ref{fig:snippet_modeling}(b) is used as the basic component to construct the backbone for GaitSnippet, where the smoothing layer is implemented using a $1 \! \times \! 1$ convolution.
The number of blocks (\emph{blocks}), convolutional channels (\emph{channels}), and strides (\emph{strides}) for each stage across the four datasets are detailed in Table~\ref{tab:settings}, referring to the network configurations used in~\cite{fan2023exploring,ma2024learning}, \eg, \emph{blocks} = $[1,4,4,1]$ for DeepGaitV2 on Gait3D~\cite{fan2023exploring} and \emph{blocks} = $[3,4,6,3]$ for VPNet on GREW~\cite{ma2024learning}.
%
%
The settings for CCPG and CCGR-MINI follow those used for Gait3D, with one modification: the backbone architecture used for CCPG employs a reduced number of \emph{blocks}, specifically $[1, 1, 1, 1]$.

Besides, the margin threshold $\delta$ for triplet loss is set to $0.2$ and the loss weight $\alpha$ for snippet-level supervision is set to $0.75$.
During inference, \emph{Snippet-level Branch} is disabled, and we use \emph{Sequence-level Branch} output for similarity computation.
The feature dimensions match DeepGaitV2 (\eg, $16 \! \times \! 256$ on Gait3D/GREW).
For other settings, such as data preprocessing and training strategies, we refer to those described in~\cite{fan2023exploring,ma2024learning}.
%
%
To ensure reproducibility, the PyTorch-based source code and pretrained models will be made publicly available.

\subsection{More Experimental Results}

\subsubsection{Analyses on Computation Cost}
\label{sec:computation_cost}
In Figure~\ref{fig:ablation_cost}, we compare the computational cost of GaitSnippet with several representative methods in terms of parameter count and FLOPs.
The statistics are obtained on the Gait3D dataset, following the methodology of~\cite{wang2023hih,huang2025occluded}.
We focus our comparison on DeepGaitV2-2D/3D/P3D, all of which adopt the same network depth on Gait3D. 

\begin{enumerate}[(1)]
	\item Notably, GaitSnippet has significantly fewer parameters and FLOPs than DeepGaitV2-3D, and even fewer than DeepGaitV2-P3D.
	
	\item Compared to DeepGaitV2-2D, GaitSnippet exhibits a slightly higher computational cost, primarily due to the introduction of smoothing layers and temporal aggregation in intra-snippet modeling.
	However, this modest increase is justified by a substantial performance improvement over DeepGaitV2-2D, with a +9.3\% gain in Rank-1 accuracy and a +9.0\% gain in mAP on Gait3D.
	It is noteworthy that GaitSnippet also outperforms both DeepGaitV2-3D and DeepGaitV2-P3D.
	
	\item To rule out the impact of model size, we further evaluate a lightweight version of GaitSnippet with reduced network depth (\emph{blocks} = $[1,3,3,1]$), resulting in only 22.8M parameters and 68.6G FLOPs lower than those of DeepGaitV2-2D.
	Despite this compact design, the model achieves competitive performance (Rank-1: 77.0\%, mAP: 69.3\%), still outperforming DeepGaitV2-2D and even DeepGaitV2-3D/P3D.
	This confirms that the performance gains stem from the proposed snippet-based modeling rather than increased model complexity.
\end{enumerate}

\subsubsection{Ablation Study on Evaluation}
During the evaluation phase, we fix $L_1 \! = \! 16$, which requires only a single forward pass and does not add additional inference burden.
In this section, we explore different values of $L_1$ (\eg, $L_1 \! = \! 8$) and evaluate the ensemble of multiple segment partitions (\eg, $L_1 \! \in \! \{8, 16\}$) \emph{only for evaluation}.
The results in Table~\ref{tab:ablation_evaluation} indicate that:
(1) The segment partition has a minor effect on evaluation, as the model is trained with various partition strategies.
(2) Averaging the features from multiple segment partitions can slightly improve recognition performance in terms of mAP, but this significantly increases the inference cost.
Further discussion is provided in Section~\ref{sec:discuss_modeling}.

Additionally, in the last row of Table~\ref{tab:ablation_evaluation}, we experiment with using snippet-level representations for evaluation.
Snippet-level features from a sequence are averaged to match the probe and gallery. Unsurprisingly, the results are inferior to those using sequence-level representations which effectively capture long-range temporal dependencies.

\subsubsection{Ablation Study on Generalization}
In this section, we aim to verify the generalization capability of our approach by applying the snippet paradigm to some set-based methods.
It is important to note that sequence-based methods rely on \emph{continuous} input, whereas both the snippets within a sequence and the frames within each snippet are \emph{discontinuous}, making it infeasible to transform sequence-based methods into their snippet-based counterparts.
Thus, we adopt the set-based methods GaitSet~\cite{chao2019gaitset} and GaitBase~\cite{fan2023opengait} for our analysis.
Specifically:
(1) We redesign the input phase using the snippet-based sampling strategy.
(2) We insert a Snippet Block, as illustrated in Figure~\ref{fig:snippet_modeling}, between two convolutional layers in each stage of the backbone.
Finally, we observe significant performance gains, as shown in Table~\ref{tab:ablation_backbone}.

\begin{table}[!tbp]
	\setlength{\tabcolsep}{10pt}
	\begin{center}
		\resizebox{0.75\linewidth}{!}{%
			\begin{tabular}{c|c|cc}
				\hline
				Feature Type & $L_1$ (\emph{for evaluation}) & R1 & mAP \\
				\hline
				\emph{Sequence-level Features} & 16 & 77.5 & 69.4 \\
				\emph{Sequence-level Features} & 8 & 77.3 & 69.5 \\
				\emph{Sequence-level Features} & \{8, 16\} & 77.2 & 69.8 \\
				\hline
				\emph{Snippet-level Features} & 16 & 65.2 & 54.8 \\
				\hline
			\end{tabular}
		}
	\end{center}
	\caption{
		Ablation study on evaluation.
		\emph{Sequence-level Features} and \emph{Snippet-level Features} denote the features before BNNeck generated by the sequence-level and snippet-level branches, respectively.
		$L_1$ denotes the length of the first segment \emph{for evaluation}.
		For the ensemble of different $L_1$ values, we average the features from multiple sequence partitions.
		The results are reported on Gait3D in terms of rank-1 accuracy (R1, \%) and mean Average Precision (mAP, \%).
	}
	\label{tab:ablation_evaluation}
\end{table}
\begin{table}[!tbp]
	\setlength{\tabcolsep}{20pt}
	\begin{center}
		\resizebox{0.75\linewidth}{!}{%
			\begin{tabular}{c|cc}
				\hline
				Model & R1 & mAP \\
				\hline
				GaitSet~\cite{chao2019gaitset} & 36.7 & 30.0 \\
				GaitSet~\cite{chao2019gaitset} + Snippet & \bftab{48.2} & \bftab{39.1} \\
				GaitBase~\cite{fan2020gaitpart} & 64.6 & 55.3 \\
				GaitBase~\cite{fan2020gaitpart} + Snippet & \bftab{69.7} & \bftab{60.2} \\
				\hline
			\end{tabular}
		}
	\end{center}
	\caption{
		Ablation study on generalization.
		Experiments are conducted on Gait3D, with results reported in terms of rank-1 accuracy (R1, \%) and mean Average Precision (mAP, \%).
	}
	\label{tab:ablation_backbone}
\vspace{-2ex}
\end{table}

\subsubsection{Ablation Study on Frame Sampling}
In Section~\ref{sec:snippet_sampling}, we provide a brief review of the sampling strategies used in recent set-based and sequence-based studies~\cite{chao2019gaitset,fan2020gaitpart,lin2021gait}.
Typically, a limited number of frames (\ie, $S\!=\!30$ in most cases) are sampled during training for each sequence.
In our experiments, however, we sample $S\!=\!32$ frames per sequence, corresponding to $M\!=\!4$ snippets per sequence and $N\!=\!8$ frames per snippet.
For a rigorous ablation study, we also apply a sampling strategy of $S\!=\!32$ frames per sequence to several representative baseline methods~\cite{fan2023exploring}, which has a slight impact on performance, as indicated in Table~\ref{tab:ablation_sampling_frames}.

\subsubsection{Snippets on Missing Frames}
In GaitSnippet, we partition each trajectory into contiguous segments and randomly sample frames within each segment to form snippets. This reduces reliance on ``perfect'' gait cycles and emphasizes recurring cues that are robust to frame dropping.

To further validate this, we simulate different frame-dropping ratios at test time on Gait3D and report rank-1 accuracy in Table~\ref{tab:ablation_missing_frames}.
GaitSnippet consistently outperforms DeepGaitV2-P3D, and the margin increases as fewer frames are retained.

\begin{table}[!tbp]
	\setlength{\tabcolsep}{4pt}
	\begin{center}
		\resizebox{0.8\linewidth}{!}{%
			\begin{tabular}{c|c|c|cc}
				\hline
				Model & \tabincell{c}{Sampling\\Strategy} & $S$ & R1 & mAP \\
				\hline
				DeepGaitV2-2D~\cite{fan2023exploring}     & \emph{Set} & $\mathrm{min}\!:\!10,\mathrm{max}\!:\!50$  & 68.2 & 60.4 \\
				DeepGaitV2-3D~\cite{fan2023exploring}     & \emph{Seq} & 30  & 72.8 & 63.9 \\
				DeepGaitV2-P3D~\cite{fan2023exploring}    & \emph{Seq} & 30  & 74.4 & 65.8 \\
				\hline
				\hline
				DeepGaitV2-2D~\cite{fan2023exploring}     & \emph{Set} & 32  & 68.6 & 60.1 \\
				DeepGaitV2-3D~\cite{fan2023exploring}     & \emph{Seq} & 32  & 72.1 & 64.6 \\
				DeepGaitV2-P3D~\cite{fan2023exploring}    & \emph{Seq} & 32  & 74.2 & 65.8 \\
				\hline
			\end{tabular}
		}
	\end{center}
	\caption{
		Ablation study on frame sampling.
		\emph{Set} and \emph{Seq} represent the set-based~\cite{chao2019gaitset} and sequence-based~\cite{fan2020gaitpart} sampling strategies, respectively.
		$S$ denotes the number of frames sampled per sequence during the training phase.
		In the original implementation of DeepGaitV2-2D~\cite{fan2023exploring}, $S$ is randomly selected from $\{10, 11, \cdots, 49, 50\}$.
		We conduct the experiments on Gait3D and report the results in terms of rank-1 accuracy (R1, \%) and mean Average Precision (mAP, \%).
	}
	\label{tab:ablation_sampling_frames}
\vspace{-2ex}
\end{table}

\subsubsection{Snippets on Skeleton Maps}
Our snippet paradigm can also be applied to skeleton maps~\cite{fan2024skeletongait}. We conduct additional experiments by directly transferring snippet-based modeling to the skeleton-map setting, as shown in Table~\ref{tab:ablation_skeleton_maps}:
\begin{enumerate}[(1)]
    \item \textbf{GaitSnippet (SkeletonMap)}: we simply replace silhouette inputs with skeleton maps, while keeping the snippet sampling strategy and Snippet Blocks unchanged.
    \item \textbf{GaitSnippet (Silhouettes+SkeletonMap)}: a multi-modal extension that follows the fusion strategy of SkeletonGait++, but replaces both the silhouette and skeleton branches with our snippet-based counterparts.
\end{enumerate}

In these experiments, both variants consistently outperform SkeletonGait and SkeletonGait++, respectively. This empirically demonstrates that i) the snippet paradigm is \emph{not tied to silhouettes} but serves as a general temporal modeling scheme that can be plugged into skeleton-map-based architectures; and ii) even in the multi-modal setting, snippet-based modeling brings further gains.

\subsection{More Discussion}
\label{sec:discussion}

\subsubsection{Discussion on Snippet Sampling}
\label{sec:discuss_sampling}
In this section, we provide additional clarifications on Snippet Sampling from three perspectives:

(1) \emph{Frame Order for Segment Partition}.
As clarified in Section~\ref{sec:snippet_sampling}, the snippet-based sampling relies on frame order to partition the sequence into segments.
As a result, the snippet-based modeling is \emph{\textbf{not}} permutation invariant to frame order, making it inappropriate to categorize it into the set-based category~\cite{zaheer2017deep}.
In our formulation, frames in a snippet are randomly sampled from a continuous segment of the sequence to describe an action, which forms the basis for crucial local context modeling.

(2) \emph{Random Selection within Each Segment}.
Randomly selecting frames within each segment for snippet sampling reduces dependency on continuous input and enhances robustness to missing silhouettes.
An interesting future direction could involve sampling snippets by identifying important silhouettes~\cite{hou2022gait,wang2024qagait}; however, measuring frame importance is challenging, especially during the input phase.

(3) \emph{Data Augmentation}.
The augmentation strategy of varying the first segment to increase snippet diversity is naturally suited for snippet-based gait recognition.
Similarly, state-of-the-art set-based and sequence-based methods adopt their own specific augmentation strategies, such as randomly sampling frames from a continuous segment in sequence-based methods~\cite{fan2020gaitpart,wang2023hierarchical}.
As evidenced in previous works, such augmentation does not hinder fair comparisons under the same evaluation protocol~\cite{hou2022comprehensive}.

(4) \emph{Sampling Hyper-parameters}.
During training, \emph{sampling with replacement} is adopted when the sequence length is insufficient.
If the number of segments $K$ is smaller than $M$, some segments may be sampled multiple times.
Similarly, if a segment contains fewer than $N$ frames, repeated frames may appear within a snippet.
During inference, all segments are used ($M=K$), and each snippet includes all frames in the segment ($N=L$).

It is worth noting that, the segment length $L$ is an important hyper-parameter for both training and inference, and $K$ is derived from the sequence length and $L$.
We set $L\!=\!16$ as an empirical approximation of the frames per gait cycle, based on previous study~\cite{ma2024learning}, dataset analysis~\cite{zheng2022gait,zhu2021gait}, and the ablation studies in Table~\ref{tab:ablation_sampling}.
Although gait cycles vary across individuals and walking speeds, $L\!=\!16$ serves as a reasonable average~\cite{ma2024learning}.
%
%
Further experiments on CCGR-MINI shown in Table~\ref{tab:comparison_emerging}, which covers diverse walking speeds, reaffirm the model's effectiveness under varying temporal conditions.
We will explore dynamic estimation of gait cycles for improved adaptability.

\begin{table}[!tbp]
	\setlength{\tabcolsep}{10pt}
	\begin{center}
		\resizebox{0.75\linewidth}{!}{%
			\begin{tabular}{c|cccc}
				\hline
                Reserved Frame Ratio & 1/1 & 1/2 & 1/3 & 1/4 \\
                \hline
                DeepGaitV2-P3D~\cite{fan2023exploring} & 74.4 & 69.6 & 63.5 & 57.1 \\
                GaitSnippet    & \bftab{77.5} & \bftab{73.7} & \bftab{70.3} & \bftab{66.2} \\
                Improvement    & \bftab{+3.1} & \bftab{+4.1} & \bftab{+6.8} & \bftab{+9.1} \\
				\hline
			\end{tabular}
		}
	\end{center}
	\caption{
		Snippets on missing frames.
		We conduct experiments on Gait3D and report rank-1 accuracy (R1, \%).
	}
	\label{tab:ablation_missing_frames}
\end{table}
\begin{table}[!tbp]
	\setlength{\tabcolsep}{20pt}
	\begin{center}
		\resizebox{0.75\linewidth}{!}{%
			\begin{tabular}{c|cc}
				\hline
                Method & R1 & mAP \\
                \hline
                SkeletonGait~\cite{fan2024skeletongait} & 38.1 & 28.9 \\
                GaitSnippet (SkeletonMap) & \bftab{40.3} & \bftab{31.0} \\
                \hline
                SkeletonGait++~\cite{fan2024skeletongait} & 77.6 & 70.3 \\
                GaitSnippet (Silhouettes+SkeletonMap) & \bftab{78.8} & \bftab{73.6} \\
				\hline
			\end{tabular}
		}
	\end{center}
	\caption{
		Snippets on skeleton maps.
        Experiments are conducted on Gait3D, with results reported in terms of rank-1 accuracy (R1, \%) and mean Average Precision (mAP, \%).
	}
	\label{tab:ablation_skeleton_maps}
\vspace{-2ex}
\end{table}

\subsubsection{Discussion on Snippet Modeling}
\label{sec:discuss_modeling}
In this section, we further compare GaitSnippet with the set-based and sequence-based modeling. Additionally, we analyze the role of Temporal Max Pooling in our framework and discuss potential directions for improvement.

(1) \emph{Comparisons to Set-based Modeling}.
The potential to exploit short-term context is a fundamental advantage of the snippet-based paradigm compared to set-based methods.
Our framework focuses on leveraging short-term context modeling to enhance frame-level feature extraction and differs from set-based methods~\cite{chao2019gaitset,fan2023opengait} in four distinct ways:
i) \emph{Sampling}: Snippet sampling \emph{relies on frame order} for sequence partition and \emph{constructs two hierarchical sets} for sampled frames.
The outputs are \emph{\textbf{not}} permutation-invariant to frame order as clarified above.
ii) \emph{Modeling}: We propose an efficient and effective Snippet Block for local context modeling, integrated between two spatial convolutions in a residual block to assist in frame-level feature extraction.
iii) \emph{Supervision}: Fine-grained snippet-level supervision is introduced to further enhance the training process.
iv) \emph{Performance}: With a backbone composed of 2D convolutions, GaitSnippet significantly outperforms the best set-based method, DeepGaitV2-2D, by a large margin (\eg, R1-+9.3\% and mAP-+9.0\% on Gait3D).

(2) \emph{Comparisons to Sequence-based Modeling}.
The potential to capture long-term dependencies is another fundamental advantage of the snippet-based paradigm compared to sequence-based methods, as \emph{snippets sampled from a sequence are not temporally continuous and likely cover long-term frames}.
As a pioneering attempt, GaitSnippet improves long-term modeling through two key aspects:
i) \emph{Fine-grained Snippet-level Supervision}: This encourages the features of snippets from the same sequence to remain similar, even when there is a large temporal interval between snippets.
ii) \emph{Diverse Sequence-level Representations}: Unlike sequence-based methods, the input for Temporal Max Pooling at the end of the backbone to derive sequence-level representations changes from continuous frames to snippets that span long-term frames.
It significantly enhances the diversity of sequence-level representations for training, which helps the recognition head (\ie, HPM and BNNeck) better adapt to long-term modeling.
%

(3) \emph{Role of Temporal Max Pooling}:
i) Temporal Max Pooling is indeed an effective manner for feature aggregation in unordered sets~\cite{zaheer2017deep,chao2019gaitset,fan2023opengait}.
Taking GaitSet~\cite{chao2019gaitset} as an example, it applies it to all frames per sequence, treating them as a single set. In contrast, GaitSnippet organizes data \emph{\textbf{hierarchically}}, \ie, frames per snippet and snippets per sequence, enabling more structured modeling.
ii) GaitSet adopts Temporal Max Pooling \emph{after the backbone} for global aggregation, whereas GaitSnippet integrates it \emph{within the backbone} to enhance frame-level feature extraction via snippet-level context.
iii) In our design, Temporal Max Pooling serves as a pioneering component for intra-snippet gathering, consistently yielding performance gains across benchmarks.
We will explore more advanced hierarchical temporal aggregation strategies under the snippet paradigm.

(4) \emph{Limitations and Further Improvement}. It is important to acknowledge that our solution for snippet-based gait recognition is not necessarily optimal, and we recognize some limitations in GaitSnippet.
For example, in the inference phase, while we achieve superior performance with a single forward process (\ie, $L_{1} \! = \! 16$), multiple forward processes are required to benefit from different partition strategies (\ie, $L_{1} \! \in \! \{8,16\}$).
Incorporating partition ensemble into a single forward process represents a meaningful direction for future work.
Despite this, our solution consistently shows performance gains across various benchmarks, suggesting that snippet-based gait recognition is a highly promising approach.

\subsection{More Literature Review}
In Section~\ref{sec:related}, we primarily review silhouette-based gait recognition methods, which fall under the appearance-based category.
For completeness, we also provide a brief summary of model-based gait recognition here.

Model-based approaches primarily focus on explicitly modeling the walking process.
While early research in this category relied on hand-crafted features~\cite{lee2002gait,zhang2007human}, recent studies predominantly leverage 2D/3D pose representations~\cite{liao2020model,teepe2021gaitgraph,teepe2022towards,zhang2023spatial,pinyoanuntapong2023gaitmixer,guo2023physics,fan2024skeletongait} or SMPL parameters~\cite{li2020end,li2022multi} as input for data-driven feature learning using deep neural networks.
For instance, Teepe~\etal~\cite{teepe2021gaitgraph} model 2D poses as graphs and process pose sequences using a Graph Convolutional Network.
Fu~\etal~\cite{fu2023gpgait} enhance the generalization capability of 2D pose-based gait recognition by applying normalization techniques to the input and extracting fine-grained features.
Guo~\etal~\cite{guo2023physics} develop a physics-augmented auto-encoder framework for 3D pose-based gait recognition.
SkeletonGait~\cite{fan2024skeletongait} converts 2D pose data into a heatmap-like representation, enabling the use of Convolutional Neural Networks for feature extraction.

Moreover, emerging research has focused on fusing multiple modalities~\cite{hsu2022gaittake,peng2023learning,cui2023multi} for gait recognition or exploring new modalities, including RGB images~\cite{liang2022gaitedge,ye2024biggait}, point clouds~\cite{shen2023lidargait,han2022licamgait}, and event cameras~\cite{wang2019ev}.

\subsection{The Use of Large Language Models (LLMs)}
In preparing this manuscript, Large Language Models (LLMs) were used solely for checking potential grammatical and stylistic issues in the writing.
The use of LLMs did not influence the development of research ideas, experimental design, data analysis, or the interpretation of results.
All scientific contributions, including the methodology, experiments, and conclusions, are entirely the work of the authors. 